\documentclass[letterpaper, 10 pt, journal, twoside]{ieeetran}

\usepackage{amsmath,amsfonts}
\usepackage{algorithmic}
\usepackage{algorithm}
\usepackage{array}
\usepackage[caption=false,font=normalsize,labelfont=sf,textfont=sf]{subfig}
\usepackage{textcomp}
\usepackage{stfloats}
\usepackage{url}
\usepackage{verbatim}
\usepackage{graphicx}
\usepackage{cite}
\usepackage{layouts}
\usepackage{bm}
\usepackage{caption}
\usepackage{subcaption}
\usepackage{csvsimple}
\usepackage{pgfplotstable}
\usepackage{afterpage}
\usepackage{xcolor}

\begin{document}

    \begin{minipage}{0.5\textwidth} 
    \large
        IEEE Copyright Notice\\~\\
        \normalsize
        \setlength{\parindent}{1em} 
        \fbox{\parbox{\linewidth}{ 
            © 2024 IEEE.  Personal use of this material is permitted.  Permission from IEEE must be obtained for all other uses, in any current or future media, including reprinting/republishing this material for advertising or promotional purposes, creating new collective works, for resale or redistribution to servers or lists, or reuse of any copyrighted component of this work in other works.\\~\\

            Accepted to be published in: IEEE Robotics and Automation Letters
(RA-L), 2024.\\~\\

DOI: 10.1109/LRA.2024.3399590
        }}
    \end{minipage}

\clearpage 
\newpage


\title{Enhanced Model-Free Dynamic State Estimation for a Soft Robot Finger Using an Embedded Optical Waveguide Sensor}

\author{Henrik Krauss$^{1}$ and Kenjiro Takemura$^{2}$, \IEEEmembership{Member, IEEE}%

\thanks{Manuscript received: December 18, 2023; Revised March 20, 2024; Accepted May 5, 2024.}
\thanks{This paper was recommended for publication by Editor Y.-L. Park upon evaluation of the Associate Editor and Reviewers' comments.} 
\thanks{$^{1}$ Henrik Krauss is with the Graduate School of Science and Technology, Keio University, Yokohama 223-8522, Japan
{\tt\footnotesize henrik1.krauss@gmail.com}}
\thanks{$^{2}$ Kenjiro Takemura is with the Department of Mechanical Engineering, Keio University, Yokohama 223-8522, Japan
{\tt\footnotesize takemura@mech.keio.ac.jp}}
\thanks{Digital Object Identifier (DOI): 10.1109/LRA.2024.3399590}
}
\markboth{IEEE Robotics and Automation Letters. Preprint Version. Accepted May 2024}
{Krauss and Takemura: Model-Free State Estimation for a Soft Robot Finger Using an Embedded Waveguide Sensor} 

\maketitle

\begin{abstract}
In this letter, an advanced stretchable optical waveguide sensor is implemented into a multidirectional PneuNet soft actuator to enhance dynamic state estimation through a NARX neural network. The stretchable waveguide featuring a semidivided core design from previous work is sensitive to multiple strain modes. It is integrated into a soft finger actuator with two pressure chambers that replicates human finger motions. The soft finger, designed for applications in soft robotic grippers or hands, is viewed in isolation under pneumatic actuation controlled by motorized linear stages.
The research first characterizes the soft finger's workspace and sensor response. Subsequently, three dynamic state estimators are developed using NARX architecture, differing in the degree of incorporating the optical waveguide sensor response. Evaluation on a testing path reveals that the full sensor response significantly improves end effector position estimation, reducing mean error by 51\% from 5.70 mm to 2.80 mm, compared to 
only 21\% improvement to 4.53 mm using the estimator representing a single core waveguide design. The letter concludes by discussing the application of these estimators for (open-loop) model-predictive control and recommends future focus on advanced, structured soft (optical) sensors for model-free state estimation and control of soft robots.
\end{abstract}


\section{Introduction}
\label{sec:introduction}

\IEEEPARstart{I}{n} contrast to their rigid counterparts, soft robots require internal sensing to achieve accurate proprioception, meaning the ability to sense their own movement and pose. However, precise state estimation and control in soft robots remains challenging due to high nonlinearities, their inherent underactuation, hysteresis in material behavior, and changes in behavior over time \cite{wang_perceptive_2018,li_soft_2022,dellasantina_modelbased_2023}.
These characteristics increase the difficulty for analytical and numerical modelling for control of soft robots especially in the highly dynamic case. Examples of modelling methods used are the constant curvature model, the Cosserat rod theory, or Finite Element Method (FEM)-based models \cite{schegg_review_2022,armanini_soft_2023}.

Model-free approaches can provide one viable alternative.
Here, representations of the robot statics or dynamics is created through machine learning on pre-recorded data \cite{thuruthel_learning_2017,georgethuruthel_control_2018,bhagat_deep_2019,kim_review_2021}.
Further, a model-free representation allows for a direct implementation of the sensor response without the need of characterizing the nonlinear behaviour within a complex soft robot system manually \cite{thuruthel_soft_2019,truby_distributed_2020,georgethuruthel_closing_2022}. This is especially interesting for multimodal (soft) sensors where usually manual signal decoupling methods are applied \cite{yang_multimodal_2022}.

One category of embedded soft sensors are stretchable or soft optical waveguide sensors, hereinafter referred to as SOWS. Classical SOWS feature a single round or rectangular optical light core and have been developed using various manufacturing techniques for applications in soft robotics, biomedicine or as wearable sensors \cite{wu_soft_2021,wang_soft_2023}.
One approach to increase sensing capabilities of SOWS is to increase the number of individually embedded waveguides \cite{zhao_optoelectronically_2016,teeple_soft_2018,shen_soft_2021}. For optical waveguides, optical effects and structural light guiding can and should be used to enhance sensing capability.
Recently, more advanced SOWS have been developed that can distinguish multiple degrees of freedom or enable multimodal sensing such as by implementing dyed regions in the waveguide core to distinguish strain position \cite{bai_stretchable_2020,kim_heterogeneous_2020,sun_datadriven_2023}.
In our previous work, a structurally advanced stretchable optical waveguide featuring a semidivided core cross section has been developed that is sensitive to multiple strain modes, i.e. elongation, bending, twisting, local pressure \cite{krauss_stretchable_2022,krauss_design_2022}.
However, SOWS are hard to create a sensor model for due to manufacturing differences and even more so in the advanced case it is hard to decouple individual strain modes.
Here as mentioned before, machine learning can be used for decoupling multimodal SOWS \cite{sun_datadriven_2023, kim_heterogeneous_2020}. There is potential for using advanced, multimodal SOWS for model-free state estimation and control of soft actuators which shall be explored in this study.

Developing multidirectional soft finger actuation holds significance in the context of soft robotic or prosthetic hands, aiming to emulate the natural movements of the human hand and thumb motions \cite{piazza_century_2019,mendez_current_2021}. This extends to soft wearable gloves and devices, enhancing their capability to support flexion, abduction, and adduction motions effectively, as highlighted in \cite{ge_design_2020,yin_wearable_2021,zhu_soft_2022}. Additionally, multidirectional actuators find application in the advancement of soft grippers for control of sideways actuator movement \cite{zhou_bioinspired_2021}.

The purpose of this study is to showcase the effectiveness of structured, multimodal SOWS in enhancing the dynamic state estimation of soft robots using machine learning techniques only, without the need of decoupling the strain modes from the sensor signal.
To achieve this, a multimodal stretchable optical waveguide sensor is implemented into a pneumatic soft actuator capable of multidirectional motion. The dynamic state estimation is established using a recurrent neural network.

\section{Methods}
\subsection{Optical waveguide-integrated PneuNet soft actuator}
\begin{figure*}[!t]
\begin{center}
\includegraphics[width=\textwidth]{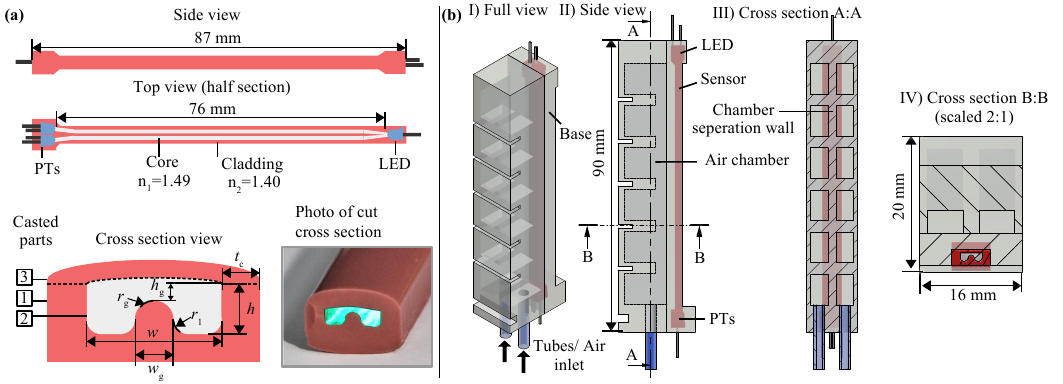}
\caption{Custom PneuNet finger with integrated waveguide featuring a semidivided core. (a) Waveguide design and cross section photo. (b) Four views of the full PneuNet finger design.}
\label{fig_PN_Setup}
\end{center}
\end{figure*}
\begin{figure*}[t]
\begin{center}
\includegraphics[width=\textwidth]{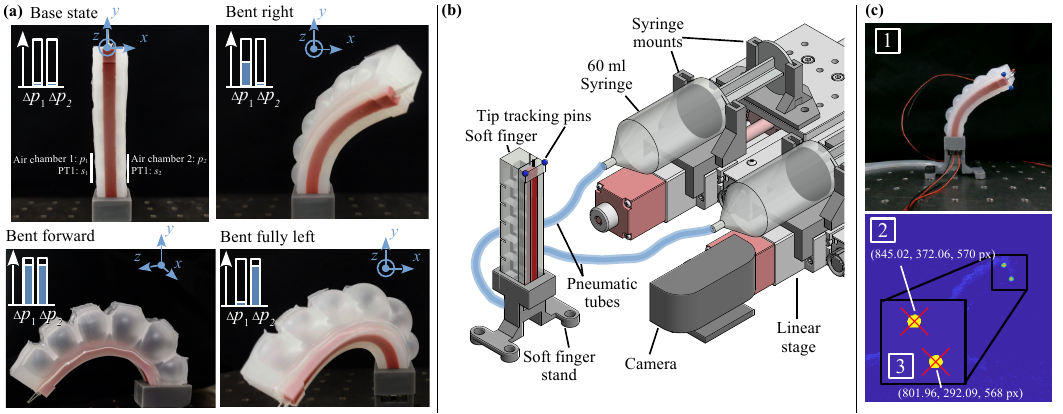}
\caption{2DOF PneuNet soft finger actuation and system. (a) Example actuation states including full pressure in both inlets (flexion) as well as left and right bending for the PneuNet finger. (b) Pneumatic system for soft finger actuation and image capturing setup. (c) Example image of the recorded data set and its image processing.}
\label{fig_PN_Actuation_System}
\end{center}
\end{figure*}
A stretchable optical waveguide sensor is implemented into a PneuNet soft actuator. The waveguide design is illustrated in Fig. \ref{fig_PN_Setup}(a) and the cross-section dimensions are given in table \ref{tab_waveguide_dim}. The waveguide sensor features a semidivided core that corresponds to an adjusted design of previous work \cite{krauss_stretchable_2022}.
The core cross section has been flattened and widened and the gap width has been increased.
The first casted part of the cladding as numbered in Fig. \ref{fig_PN_Setup}(a) does not include an overhang in the outer walls. The top layer of the core is then rounded by surface tension of the uncured resin. This leads to lower light attenuation compared to the previous design and more consistent manufacturing results.

The three casting manufacturing steps are performed using the silicone rubber Elastosil M4601 by Wacker Chemie for the cladding and the transparent urethane rubber Clear Flex 30 by Smooth-on yielding a higher refractive index as the core material. As the light source, one Kingbright $880$ nm near-infrared LED is used and and two Osram SFH 309 FA near-infrared phototransistors are used as light detectors. They exhibit a maximum sensitivity at 900 nm. The total sensor length is $87$ mm. 

The main working principle is expected to remain the same. The connecting top layer between both semidivided cores causes a light interchange which can be asymmetric under deformation and result in different signal strengths obtained at both light detectors. This enables sensitivity toward elongation, local deformation amplitude and position, twisting amplitude and direction, as well as bending amplitude and direction.

A custom PneuNet finger is chosen as the soft actuator platform. Fig. \ref{fig_PN_Setup}(b) shows the custom PneuNet finger design with the sensor integrated into the base. The finger is manufactured out of the silicone rubber Smooth-on Ecoflex 00-30 which has a lower Young's modulus than Elastosil M4601 allowing the sensor to take the role of a strain-limiting layer. The internal air chamber row is separated 
  so that the two degrees of freedom in actuation allow for simulation of finger flexion as well as abduction and adduction motion. The slits separating the sub-chambers are chosen to be short to balance the ratio of flexion and sideways motion. This design constitutes a mix of a fast and slow PneuNet actuator after Mosadegh et al.
 \cite{mosadegh_pneumatic_2014}.
Two tubes are inserted into the finger base as air inlets for pneumatic actuation and the base geometry features a notch and an outdent for firm mounting.
The finger is manufactured by multi-step moulding process out of the silicone rubber Smooth-On Ecoflex 00-30 analogous to \cite{mosadegh_pneumatic_2014}. The sensor is embedded through overmoulding in the base layer.
\begin{table}[h!]
\centering
\caption{Main dimensions of the core cross section in the optical waveguide sensor.} 
\label{tab_waveguide_dim}
 \small
 \begin{tabular}{p{20mm}p{20mm}p{28mm}ll}
 
  \hline
	Dimension	&Value &Description \\\hline \hline
	$h$	&$>1.3$ mm  &Core height \\
	$h_g$	&$>0.45$ mm  &Gap height \\
	$w$	&$3.4$ mm  &Core Width \\
	$t_c$	&$1.3$ mm  &Cladding thickness \\

	$w_g$	&$1$ mm  &Gap Width \\
	$r_1$, $r_g$	&$0.4$ mm  &Fillet radius \\		
	$n_1$	&$1.47$  &Core refr. index \\
	$n_2$	&$1.39$  &Cladding refr. index \\\hline \hline
 \end{tabular}

\end{table}

\subsection{System for pneumatic actuation and image acquisition}
Fig. \ref{fig_PN_Actuation_System}(a) shows the manufactured PneuNet finger and example actuated states to show the actuator's work space range as well as its basic actuation principle. For the base state, no additional pressure is applied ($\Delta p_1=\Delta p_2 = 0$). For bending right, pressure is increased relatively in the first air chamber and vice-versa.
The thicker base layer including the waveguide sensor mitigates forward bending. This simulates the flexion motion of a human finger. The possibilities of sideways bending when unequal additional pressure is applied in both chambers simulates abduction and adduction motion.

An overview over the full system for pneumatic actuation is given in Fig. \ref{fig_PN_Actuation_System}(b). Two polymer pneumatic tubes with a total length of circa $300$ mm are connected to two $60$ mL syringes. They are placed in 3D-printed mounts on one COMS Corp. PM80B-200X motorized linear stage each.
This closed pneumatic system allows for fixed, repeatable pressure values through adjustments of the syringe plunger position. The stages are powered and controlled using a COMS Corp. CP-700 position controller which is connected to Matlab using serial data transmission over USB.

A 1080P camera recording at $25$ Hz is placed in front of the soft finger for end effector position tracking. Two blue pins of spherical shape and $2.5$ mm diameter are inserted into both ends of the soft finger tip edge. 
Matlab is used for image processing. First, the image is taken and then reduced to the blue component. After applying a median filter, the blue image is binarized using a threshold of $15 \%$ and using blob analysis the position and size in pixels of the two pins can be determined as visualized in Fig. \ref{fig_PN_Actuation_System}(c).
For conversion to positional data in mm the projection transformation model given in equation \ref{eq_cam_transf} is applied,

\begin{equation}
\begin{aligned}
r_\mathrm{mm} &= c_0 r_\mathrm{px} + c_1 r_\mathrm{px}h_\mathrm{px}, \\
z_\mathrm{mm} &= c_2 r_\mathrm{px} + c_3 h_\mathrm{px} + c_4 h_\mathrm{px}^2 + c_5,
\end{aligned}
\label{eq_cam_transf}
\end{equation}

\begin{equation}
\begin{aligned}
c_0 &= 1.9e-1, \quad c_1 = -1.99e-3, \\
c_2 &= -6.77e-3, \quad c_3 = -5.31e-2, \\
c_4 &= 7.13, \quad \phantom{-{}} c_5 = -2.87e+2
\end{aligned}
\label{eq_cam_transf_param}
\end{equation}

with $r=\sqrt{x^2+y^2}$ and $h_\mathrm{px}^2=A_\mathrm{px}$. Here, $x_\mathrm{px}$, $y_\mathrm{px}$ and $A_\mathrm{px}$ denote the pin image coordinates and pixel size and $x_\mathrm{mm}$, $y_\mathrm{mm}$ and $z_\mathrm{mm}$ real coordinates in mm. The z-axis represents the optical axis pointing toward the camera.
The six parameters $c_0$ to $c_5$ given in equation \ref{eq_cam_transf_param} are determined based on calibration measurements where blue pins are placed at $27$ positions in a $3\times3\times3$ mesh. The mesh increases in size with distance $\Delta z$ to the camera, with its width $\Delta x$ and height $\Delta y$ as given in equation \ref{eq_mesh_size}, covering the full soft finger work space.
\begin{equation}
\begin{aligned}
(\Delta z_1, \Delta z_2, \Delta z_3)=(55 \, \mathrm{mm}, 105 \, \mathrm{mm}, 155 \, \mathrm{mm}), \\
(\Delta x_1, \Delta x_2, \Delta x_3)=(100 \; \mathrm{mm}, 150 \; \mathrm{mm}, 250 \; \mathrm{mm}), \\
(\Delta y_1, \Delta y_2, \Delta y_3)=(40 \; \mathrm{mm}, 80 \; \mathrm{mm}, 120 \; \mathrm{mm})
\end{aligned}
\label{eq_mesh_size}
\end{equation}

\begin{figure*}[!t]
\begin{center}
\includegraphics[width=\textwidth]{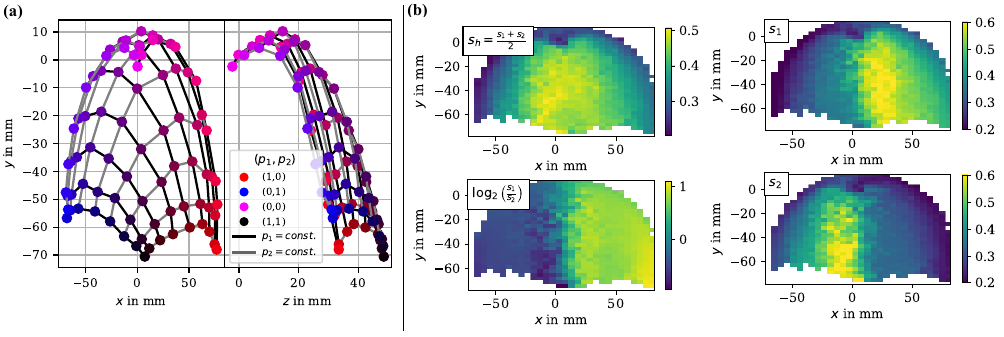}
\caption{Soft finger work space and waveguide sensor characterization. (a) Work space on a $9 \times 9$ grid of different input pressure values. (b) Waveguide sensor signal over soft finger work space for single values, average, logarithmic ratio.}
\label{fig_sens_char}
\end{center}
\end{figure*}
\subsection{Soft finger workspace}
The soft finger's work space shall be characterized. The coordinate system used is visualized in Fig. \ref{fig_PN_Setup}(a), where the origin is located at the end effector position for $p_1=p_2=0$ with the $z$-axis pointing at the camera and $x$ and $y$ denoting the horizontal and vertical axis.
Fig. \ref{fig_sens_char}(a) displays the tip end positions projected to the x-y-plane and the y-z-plane for 9-by-9 different pressure configurations. The pressure values $p_1$ and $p_2$ are linearly normalized values of the linear stages position, where $p_1=0, p_2=0$ corresponds to 60 mL of air in each syringe and ambient air pressure and $p_1=1, p_2=1$ to 25 mL of air in the syringes.
It can be seen the recorded positions span a continuous surface that is symmetric around the $y$-axis and symmetric toward the pressure values $p_1$, $p_2$. The same pressure applied in both chambers results in a pure flexion motion with $x \approx 0$ mm. When $p_1>p_2$, the soft finger bends in positive $x$-direction and vice-versa. The maximum displacement ranges observed are $\Delta x \approx 145$ mm, $\Delta y \approx 80$ mm and $\Delta z \approx 50$ mm.
\subsection{Sensor measurement and signal characterization}
In the following, the waveguide sensor response is characterized over the static soft actuator work space. The same measurement setup from previous work is used \cite{krauss_stretchable_2022}. The phototransistors' current is amplified using an LM358N operational amplifier and a main resistors of $50$ k$\Omega$.
The sensor signal strengths $s_1$, $s_2$ are linearly normalized voltages of the phototransistors placed as shown in Fig. \ref{fig_PN_Actuation_System}(a). $s_2=1$ corresponds to $5$ V and $s_2=0$ corresponds to $0$ V recorded on the analogue pin of an Arduino Mega 2560 micro-controller board. $s_1$ values are linearly scaled so that $\mathrm{max}(\bm{s_1})=\mathrm{max}(\bm{s_2})$ and $\mathrm{min}(\bm{s_1})=\mathrm{min}(\bm{s_2})$ over the sampling range described in section \ref{subsec_data_acq}.

Fig. \ref{fig_sens_char}(b) shows the signal strengths over the work space, as well as the average amplitude and ratio calculated after $\frac{s_1+s_2}{2}$ and $\log_2 (\frac{s_1}{s_2})$ respectively. It can be observed that the signal amplitude rises with higher bending amplitude. This is likely due to the top surface exhibiting lower surface roughness as it is not in contact with the 3D-printed mould. This would be expected and has been observed previously \cite{zhao_optoelectronically_2016}.
The signal measured at PT1, $s_1$ shows a significantly higher strength for pressure applied in the first chamber, namely high values of $p_1$ and vice-versa. Looking at the ratio, it can be seen that pure flexion motion $u_1=u_2$ results in a ratio of $\frac{s_1}{s_2} \approx 1$, a line separating ratios smaller than 1 for $u_2>u_1$ and larger than 1 for $u_1>u_2$. Therefore, the bending direction can be clearly distinguished and further information about the bending amplitude obtained from the waveguide sensor signal.
\section{Neural network-based dynamic state estimation}
\begin{figure*}[t]
\begin{center}
\includegraphics[width=\textwidth]{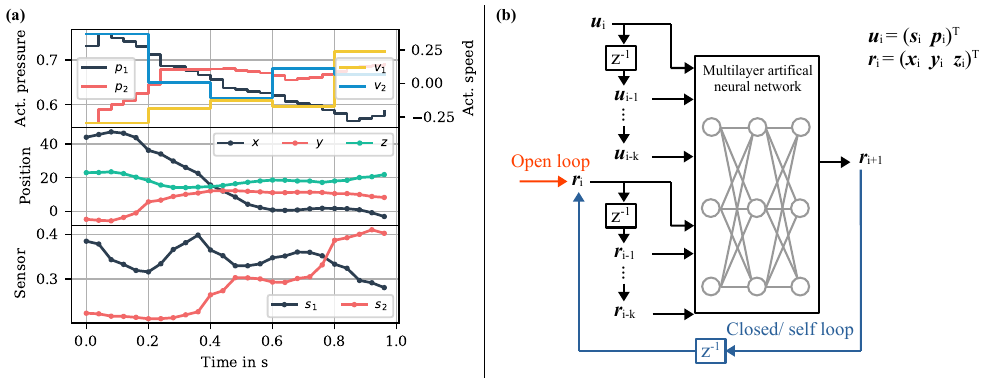}
\caption{Recorded data and neural network architecture. (a) Excerpt of actuation, recorded soft finger end effector position, and sensor data. (b) NARX neural network architecture for open and self-loop operation.}
\label{fig_exAC_NN}
\end{center}
\end{figure*}

\subsection{Data acquisition}
\label{subsec_data_acq}
The forward dynamics of the soft actuator shall be learned with a neural network. For this, the soft finger tip position and sensor response is recorded at $f_1 = 25$ Hz under random actuation.
Actuation steps occur at $f_2=5$ Hz, where the motorized linear stages are independently set at a random speed between in the intervals $\dot{p}_1/\mathrm{s}^{-1} \in [-0.4,0.4]$, $\dot{p}_2/\mathrm{s}^{-1} \in [-0.4,0.4]$ while preventing overshooting. This results in highly dynamic motion of the soft finger.
The absolute linear stage positions are then recorded at $f_1$ over serial. A total sample time of $1200$ s is chosen which is sufficient to cover the full workspace.
An excerpt of a $1$ s range of the recorded data is given in Fig. \ref{fig_exAC_NN}(a) and the full path in (b) which will be the basis for the dynamic state prediction.
\subsection{Neural network architecture and training}
A nonlinear autoregressive network with exogenous inputs (NARX) neural network architecture is used for the soft finger dynamic state prediction. The network architecture is visualized in Fig.\ref{fig_exAC_NN}(b). The soft finger tip position $\bm{r}_\mathrm{i}$ denotes the current state input as the end effector position and $\bm{u}_\mathrm{i}=(\bm{s}_\mathrm{i},\bm{p}_\mathrm{i})^\mathrm{T}$ the current exogenous inputs, including the sensor signal strength and actuation pressure values. Also, two delayed inputs are included to predict the future state after equation \ref{eq_nn},
\begin{equation}
\bm{r}_{\mathrm{i}+1}=f(\bm{r}_{\mathrm{i}},\bm{r}_{\mathrm{i}-1},\bm{r}_{\mathrm{i}-2},\bm{u}_{\mathrm{i}},\bm{u}_{\mathrm{i}-1},\bm{u}_{\mathrm{i}-2})
\label{eq_nn}
\end{equation}
where $f$ is a feedforward neural network with two hidden layers and GELU activation functions. The hidden layer sizes are determined for each model by a hyper parameter study.
After learning is completed, the network is closed and the output fed back into the state input which is referred to as self loop in contrast to closed loop operation in the context of model-predictive control. 

The network training, and prediction on the test data set is done using Python, the TensorFlow and Keras libraries, and the Adam optimization algorithm. The data is sequentially separated into sets of $90\%$ ($1080$ s) training and $10\%$ ($120$ s) testing data. 
The positional vector $\bm{r}$, the pressure values $\bm{p}$ and the sensor inputs $\bm{s}$ are normalized beforehand.
Three neural network models as given in table \ref{tab_model_configurations} are trained.
\begin{table}[ht]
\centering
\caption{Neural network model configurations}
\begin{tabular}{cccc}
\hline
Model & Sensor input & Net inputs & Hidden layers \\
\hline
\hline
$M_A$ & No signal: $\bm{s}=()$ & $15$ & $2 \times 50$ \\

$M_B$ & Averaged signal $\bm{s}=(\bm{s_h})$& $18$ &$2 \times 75$ \\
$M_C$ & Full signal: $\bm{s}=(\bm{s_1},\bm{s_2})^\mathrm{T}$ & $21$ & $2 \times 100$\\
\hline
\end{tabular}
\label{tab_model_configurations}
\end{table}
The first model $M_A$ is a state estimator solely based on the pressure input.
For the second model $M_B$ the averaged sensor signal $s_h$ as shown in Fig. \ref{fig_sens_char}(b) is used which corresponds to the signal of a single core waveguide design.
Its purpose is to assess the merit of the waveguide design with semidivided cores for the third model $M_C$. With this approach, there is no need of comparing two unique waveguide-embedded soft actuators that would be subject to manufacturing differences.
\subsection{Evaluation of results}
\label{subsec_eval_of_res}

\begin{figure}[h!]
\begin{center}
\includegraphics[]{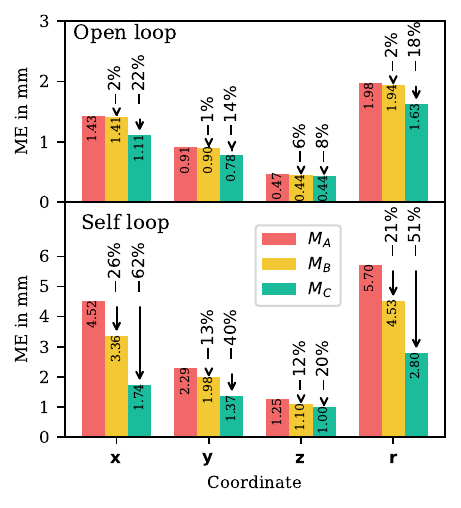}
\caption{Mean error for soft finger path estimation on the test data set in open and self loop and relative decrease of model $M_B$, $M_C$ toward $M_A$.}
\label{fig_bar_performance}
\end{center}
\end{figure}

After the NARX models are trained in open loop they are evaluated on the $120$ s test data set path. Open loop is compared to self loop prediction as visualized in Fig. \ref{fig_exAC_NN}(b). Fig.\ref{fig_bar_performance} shows the mean error ME of the distance for the coordinates and the positional vector $\bm{r}$. We observe that including the sensor information lowers the state estimation error, with model $M_C$ outperforming $M_A$ and $M_B$ in the open and self loop case. Including the waveguide sensor response shows biggest effect in the self loop, where the mean error for the soft finger end effector position $\bm{r}$ is reduced from $5.70$ mm to $2.80$ mm by $51\%$.

\begin{figure*}[!t]
\begin{center}
\includegraphics[width=0.98\textwidth]{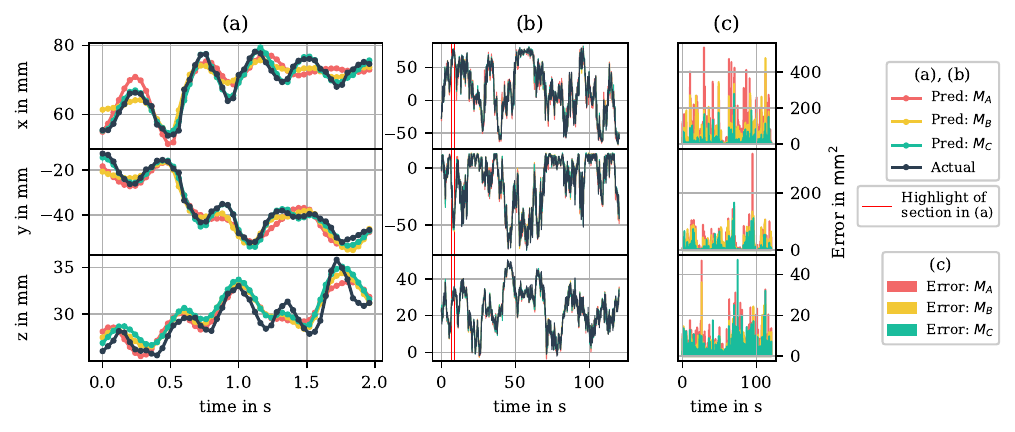}
\caption{Self-loop soft finger path estimation on test data set. (a) Highlighted path prediction section of 2 s length; (b) Full test path prediction of 120 s length including highlighting of the section shown in (a). (c) Mean square errors over full test path.}
\label{fig_xyz_path}
\end{center}
\end{figure*}

Fig. \ref{fig_xyz_path}(a) shows a highlighted region of the full soft finger path of the test data set in Fig.\ref{fig_xyz_path}(b). We can see the full $120$ s long test path in (b) is overall estimated well in continuous self loop for all models. Looking at the highlighted region in (a), the strongly improved state estimation of model $M_C$ using the full waveguide response is apparent. Especially fast oscillations for coordinates $x$ and $y$ are fully captured in comparison to model $M_A$ and $M_B$. Here, the mean error is reduced by $62\%$ and $40\%$ respectively.

A high decrease is expected for the $x$-axis as it denotes asymmetric bending of the waveguide chambers. Interestingly, the reduction of mean error for the $y$-coordinate increases threefold from the single core design ($M_B$) from $13\%$ to $40\%$ despite representing movement symmetric toward the waveguide chambers.
For coordinate $z$ we generally observe lower accuracy relative to the actual path for all models which is likely due to the tracking method of determining $\bm{z}$ from image pixel area.
Looking at the mean squared error in Fig.\ref{fig_xyz_path}(b), higher values can be seen at larger range movements. The self loop prediction is stable with no trend of an increasing MSE over the estimation horizon of $120$ s.
\section{Extending discussions}
\subsection{Application to (open-loop) model predictive control}
Having a stable long-term closed loop is required for open-loop model predictive control (OLMPC). The developed NARX models can be used for OLMPC where $M_C$ is used for past state estimation as well as one future step and $M_A$ is used for future state predictions to determine control pressure values $p_1$, $p_2$.

Due to the comparatively low frequency for self-loop
estimation at 25 Hz and low neural network sizes, an execution time 37 -- 43 times faster than real time (speedup factor) is measured on a laptop with an Intel Core i7 13620H CPU. The python code is run directly in Jupyter Notebook, using a custom NARX self-loop function.

\begin{figure}[h!]
\begin{center}
\includegraphics[]{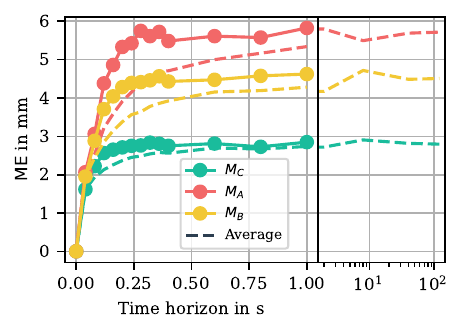}
\caption{Mean error of different time horizons for self-loop soft finger path estimation on the test data set, including the end value and average of the respective time horizons.}
\label{fig_prediction_horizon}
\end{center}
\end{figure}

Fig.\ref{fig_prediction_horizon} shows the state estimation error for models $M_A$, $M_B$, and $M_C$ for time horizons from $0.04$ s until $120$ s. We can see the ME increases until reaching a plateau after $10$ time steps corresponding to $0.4$ s. 

\subsection{Development of advanced SOWS for model-free soft robot learning}

Dynamic soft robotic system motion is challenging to estimate due its variable and multimodal character.
Stretchable sensors have the benefit of seamlessly blending into complex movements, including bending, twisting, and elongation. 
However, it is still difficult to separate these modes from the signal using a sensor model or to create a sensor that naturally aligns with the modes of the robotic system.
This latter approach has been used by Bai et al. \cite{bai_stretchable_2020} to successfully sense soft glove joint movement using advanced SOWS.

As our approach is contrary to modeling efforts, using a recurrent neural network the sensor signal does not need to be decoupled for different strain modes and a sensor model does not have to be manually created and the full state nor any physical properties of the soft robotic system need to be derived from the sensor alone. Instead it can be learned on training data individually for a soft actuator and the embedded waveguide sensor which are both strongly susceptible for manufacturing deviations, as long as the sensor signal is repeatable. In our case, increasing the number of sensor outputs by a factor of 2 improved state estimation error reduction by a factor of $2.4$.
Therefore it is recommended that future work focuses on creating more advanced, structured SOWS with multimodal sensitivity for the use in data-driven perception and proprioception models for soft robots. Here, especially optical soft sensors have the potential for increased functionality based on structural light guiding or manipulation.
\section{Conclusion}
In this study, an advanced stretchable optical waveguide sensor was implemented into the base layer of a custom PneuNet soft actuator for improving dynamic state estimation using a NARX neural network architecture.
The stretchable waveguide features a semidivided core design for sensitivity for different strain modes adjusted from previous work \cite{krauss_stretchable_2022}.
The custom pneumatic soft actuator features two pressure chambers which mimic flexion and sideways motion of human fingers. The soft finger is applicable for soft robotic grippers or hands and was considered here in isolation.

First the soft finger work space and sensor response are characterized. Then, three neural network-based dynamic state estimators ($M_A$, $M_B$, $M_C$) for the soft finger end effector positions are learned using a NARX architecture with two delayed states on a $1080$ s long highly dynamic training  at $25$ Hz. They are distinguished by not including the sensor response ($M_A$), a sensor signal averaged for both semidivided cores representing a single core design ($M_B$) and the full sensor response ($M_C$).
Evaluated on a $120$ s long testing path, the full optical waveguide sensor response significantly improved the mean error for end effector position estimation by 51\% from $5.70$ mm down to $2.80$ mm ($M_C$), compared to the signal equivalent to a single core design by 21\% down to $4.53$ mm ($M_B$).

Lastly, this letter discussed the implications of the experimental results for model-free, multimodal dynamic soft robot perception.
Future work should focus on developing more advanced, structured (optical) soft sensors with extensive multimodal sensing capability to facilitate learning-based control of soft robots.

\bibliographystyle{ieeetr}
\bibliography{bibliography}

\newpage


\end{document}